\documentclass[10pt,twocolumn,letterpaper]{article}
\pdfoutput=1

\usepackage{subfigure}
\usepackage{iccv}
\usepackage{times}
\usepackage{epsfig}
\usepackage{graphicx}
\usepackage{amsmath}
\usepackage{amssymb}
\usepackage{authblk}
\usepackage{booktabs}

\usepackage[pagebackref=true,breaklinks=true,colorlinks,bookmarks=false]{hyperref}

\iccvfinalcopy 

\def\iccvPaperID{2949} 

\ificcvfinal\fi

\begin{document}

\title{Learning to Cluster Faces via Transformer}

\author[1]{Jinxing Ye$^*$}
\author[2]{Xiaojiang Peng\thanks{Equally-contributed first authors}}
\author[1]{Baigui Sun}
\author[1,3]{Kai Wang}
\author[1]{Xiuyu Sun}
\author[1]{Hao Li \thanks{Corresponding author (lihao.lh@alibaba-inc.com)}}
\author[1]{Hanqing Wu}
\affil[1]{Alibaba Group}
\affil[2]{Shenzhen Technology University, China}
\affil[3]{National University of Singapore, Singapore}


\maketitle
\ificcvfinal\thispagestyle{empty}\fi

\begin{abstract}
Face clustering is an useful tool for applications like automatic face annotation and retrieval. The main challenge is that it is difficult to cluster images from the same identity with different face poses, occlusions, and image quality. Traditional clustering methods usually ignore the relationship between individual images and their neighbors which may contain useful context information. In this paper, we repurpose the well-known Transformer and introduce a Face Transformer for supervised face clustering. In Face Transformer, we decompose the face clustering into two steps: relation encoding and linkage predicting. Specifically, given a face image, a \textbf{relation encoder} module aggregates local context information from its neighbors and a \textbf{linkage predictor} module judges whether a pair of images belong to the same cluster or not. In the local linkage graph view, Face Transformer can generate more robust node and edge representations compared to existing methods. Experiments on both MS-Celeb-1M and DeepFashion show that our method achieves state-of-the-art performance, e.g., 91.12\% in pairwise F-score on MS-Celeb-1M.

\end{abstract}


\section{Introduction}
In recent years, face recognition has achieved remarkable progress in real-world applications due to the development of advanced metric learning methods \cite{deng2019arcface} and deep neural models \cite{2016Deep}. Large-scale well-labelled data is very crucial for high-performance face recognition system, while to annotate these large-scale datasets is time-consuming and expensive. Recent solutions resort to clustering methods aiming to mine identity information from unlabeled data.


\begin{figure}[tp]
  \centering
  \subfigure[Transfer features with context to cluster challenging samples.]{
    \label{fig:graphics:a}
    \includegraphics[scale=0.3]{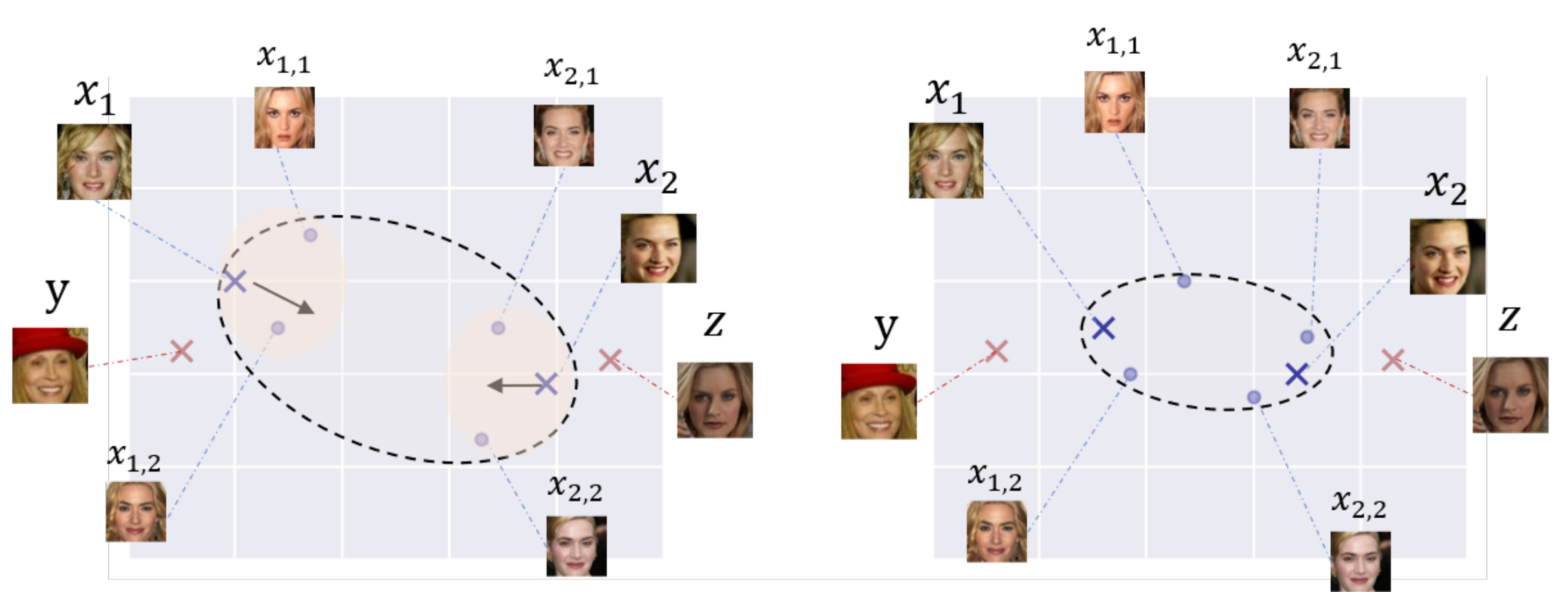}
    }
  \subfigure[Adapt edge embedding via context to ease linkage prediction.]{
    \label{fig:graphics:b}
    \includegraphics[scale=0.4]{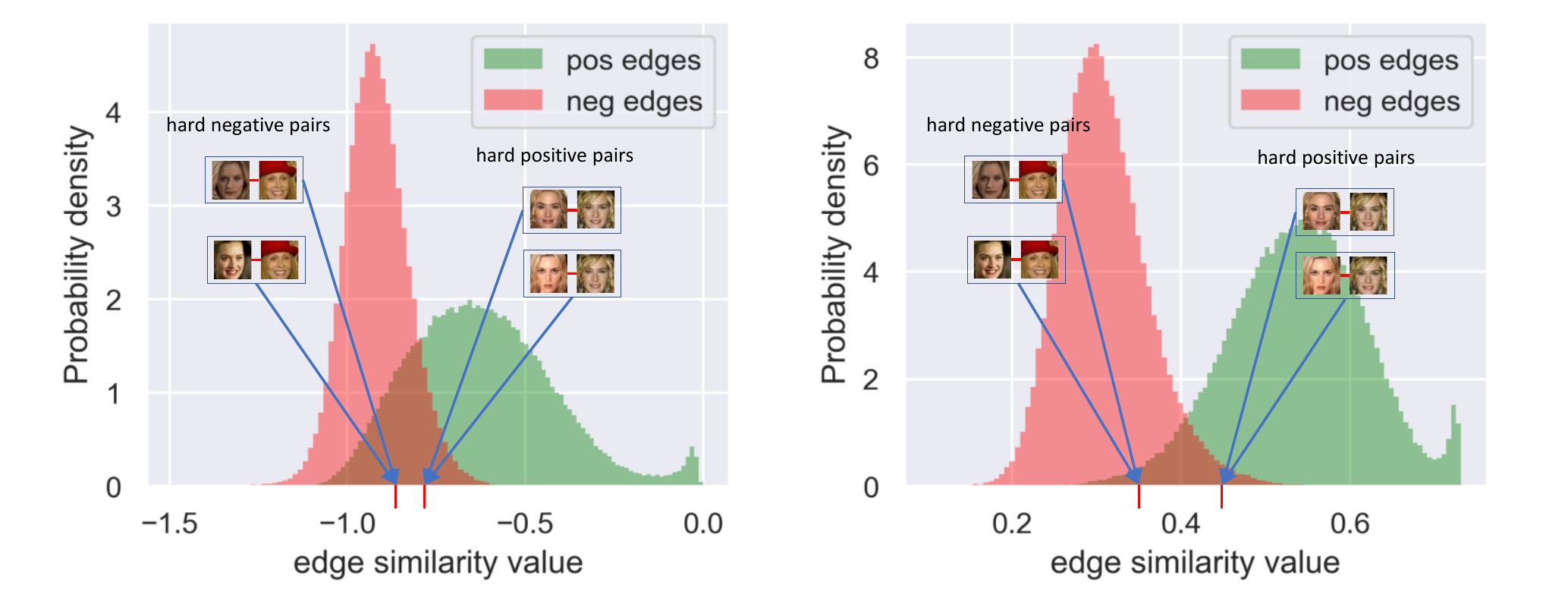}
    }
  \caption{Illustration for implementing context-based node augmentation and edge augmentation in face clustering. 
(a) Face feature distribution in wild usually exists considerable challenging samples, which is hard to present direct clustering procedure. Transferring features with its local context is helpful to cluster those challenging samples. Images from the same ID are represented by the same letters. (b) Similarity between hard positive pairs and hard negative pairs usually gathered closely, which makes it difficult to decide a threshold to distinguish whether a link could be built. Adapting edge embedding via context eases linkage prediction task.}%
  \label{fig:moti}
\end{figure}


Traditional clustering methods generally make different assumptions on the input features. K-Means \cite{lloyd1982least} requires the distribution of clusters to be convex, DBSCAN (Density-Based Spatial Clustering of Applications with Noise) \cite{ester1996density} requires the density within the cluster to be greater than a certain threshold, Spectral Clustering \cite{shi2000normalized} requires clusters are of the similar sizes. However, in real-world applications, there still exist considerable images with extreme exposure, occlusion, pose variant, and low resolution, whose distributions are often too complex to meet these distribution assumptions. We call those hard samples as challenging samples. Clustering with these challenging samples may lead to the following problems\cite{ben2014clustering}. First, challenging samples of the same person are inevitably far from the high-quality ones, resulting in splitting the same identity. Second, challenging images from different identities may be close to each other due to the dominant imaging condition, resulting in a degradation in purity. Third, the 
distance between an image and its neighbors differs from images to images, which leads to different merging threshold for different instances. The pioneer work Rank-Order \cite{zhu2011rank} and recent supervised face clustering work \cite{wang2019linkage} attempt to re-measure distance between samples via contextual information and annotations. However, they largely ignore that noise may exists in local neighbor topology, and directly consider all samples equally. 

In this paper, considering contextual information is critical for clustering on a complex distribution, we re-purpose the well-known Transformer~\cite{vaswani2017attention} and propose a uniform face clustering framework, termed as Face Transformer (FaceT). As shown in Figure \ref{fig:moti}, the key motivation of FaceT is to leverage contextual information contained in the local topology to reduce the adverse effects of challenging images thus to learn robust and compact cluster embeddings. To this end, FaceT contains two crucial modules: (i) Relation Encoder (RE) and (ii) Linkage Predictor (LP). Relation encoder is a transformer for a node to aggregate local context information from its neighbors. As illustrated in figure \ref{fig:graphics:a}, RE increases the inter-class distance and reduce the intra-class distance between samples based on their context. Linkage predictor is a transformer for a node to judge whether a neighbor belongs to the same cluster or not. As illustrated in figure \ref{fig:graphics:b}, LP improves the similarity between challenging samples and its positive samples and decreases the similarity with its negative samples.

Compared to these GCN-based supervised clustering models\cite{wang2019linkage, yang2019learning, yang2020learning, guo2020density} which need constructing adjacency matrices, our method is simpler and effectively inherits contextual information with the two transformer modules. We conduct extensive experiments with varied recognition models and training datasets, and achieve consistent improvements over related state-of-the-art methods on several widely-used benchmarks. 

Our contributions can be summarized as follows:

\begin{itemize}
\item A node enhancement structure termed Relation Encoder(RE) is proposed, which extracts the contextual information of local topical structure to enhance node embedding. 
\item A Linkage Predictor(LP) that is composed of an edge enhancement structure and linkage classifier is proposed, which regards the clustering task as a linkage prediction task and generalizes more precise predictions with enhanced edge embedding.
\item A uniformed clustering framework, \textit{i.e.} FaceT, achieves state-of-the-art performance with pairwise F-score 91.18, Bcubed F-score 90.54, NMI 97.63 on MS-Celeb-1M dataset.
\end{itemize}

\section{Related Work}
We first briefly review face clustering methods including unsupervised and supervised ones, and then present some related work on linkage prediction from social network analysis.

\subsection{Face Clustering}
\textbf{Unsupervised methods}. With the development of deep learning, recent works primarily adopt features extracted by a deep convolutional neural network (DCNN) \cite{deng2019arcface, schroff2015facenet}. For the deep feature based clustering task, traditional algorithms like K-Means, spectral clustering, and DBSCAN usually lie on different data assumptions that are difficult to satisfy. Therefore, later methods usually focus on additional contextual information to cluster faces. Rank-order\cite{zhu2011rank} proposed a relation metric approach based on the local context, ARO \cite{otto2017clustering} proposed an approximate rank-order metric to reduce rank-order running cost, DDC \cite{lin2018deep} uses minimal covering spheres of neighborhoods as the similarity metric, PHAC \cite{lin2017proximity} exploits neighborhood similarity based on linear SVMs that separates local positive instances and negative instances. Additional deep neural networks(DNNs) are recently used to boost clustering results, including unsupervised and supervised approaches. As typical unsupervised ones, DEC \cite{xie2016unsupervised} and SDLC \cite{yang2017towards} use encoder-decoder structures to learn low-dimensional embeddings and cluster assignments. In general, there have been many schemes that adjust node representation based on context and have made some progress. However, those methods depend on hand-craft information communicating policy and usually treat each node equally, making it sensitive to outliers. Therefore, such methods usually have limited performance on face clustering tasks in wild.

\textbf{Supervised methods}. As recent supervised ones, some methods use contextual information to enhance the node embedding, so as to obtain a node representation that is more friendly to clustering. VE-GCN \cite{yang2020learning} uses two stacked GCNs to estimate vertices' confidence and build edges by those high-confidence vertices, DA-NET  \cite{guo2020density} uses stacked GCNs and LSTMs to decrease class intra-distance then generate better clustering results based on traditional algorithms.  Other methods focus on the context-based distance metric method and try to obtain a more powerful distance measurement scheme. LGCN \cite{wang2019linkage} proposed a linkage-based GCN to predict the linkage between a pivot node and its neighbors via enhanced edge embeddings. Finally, some methods involve new clustering frameworks. After the clustering results are constructed using traditional methods, additional post-processing models are used to obtain more accurate clustering results. DS-GCN \cite{yang2019learning} is a typical work of this method, it learns to cluster in a detection-segmentation paradigm based on overlapped cluster proposals. Our method differs from the previous GCN-based approaches, FaceT doesn't suffer from constructing adjacency matrix and combines the advantages of the first two methods, which is more direct and effective.

\begin{figure*}[tp]
\centering
\includegraphics[width=1\textwidth]{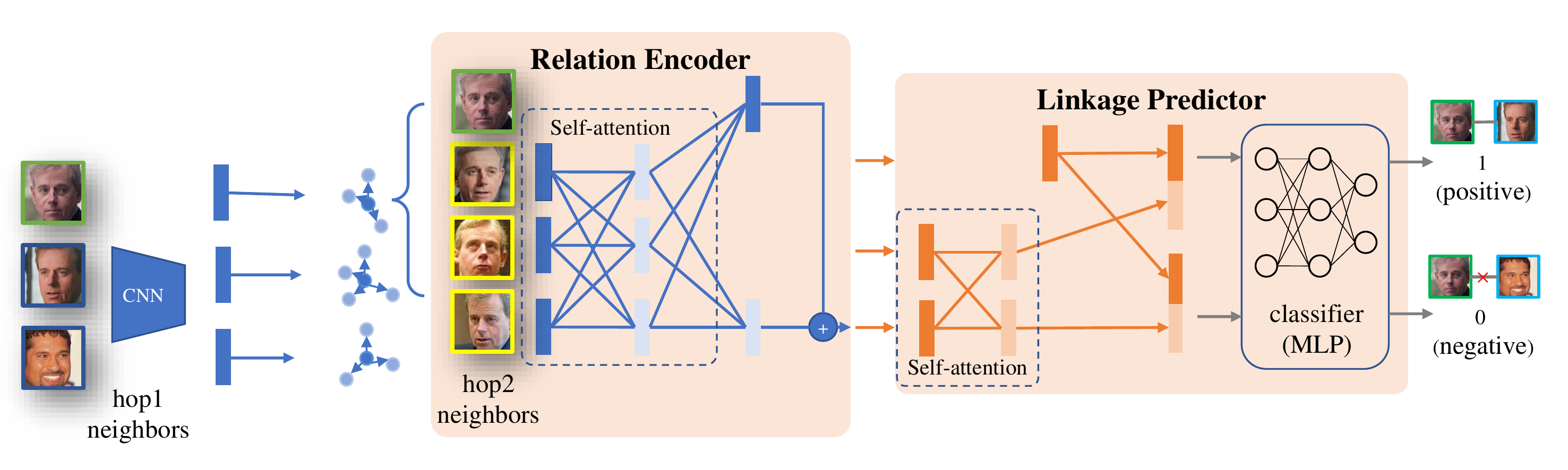}
\caption{
The overview of Face Transformer architecture. For each original face feature $f_q$, we first enhance it with its local context (constructed by $hop2$ nearest neighbors) resulting in enhanced feature $g_q$. Then for each node $f_q$, we use its $hop1$ neighbors as linkable candidates, then calculate linkage likelihood with their respective enhanced features via Linkage Predictor. Finally, we use the linkage likelihood to generate output clusters with the Union-Find algorithm. (In this toy example, $hop1$ is 2 and $hop2$ is 3.)}

\label{fig2}
\end{figure*}

\subsection{Linkage Prediction}

As a key problem in social network analysis, the objective of link prediction is to identify pairs of nodes that will either form a link or not. PageRank \cite{page1999pagerank} and SimRank \cite{jeh2002simrank} analyze the whole graph via various information propagation approaches, preferential attachment \cite{barabasi1999emergence} and resource allocation \cite{zhou2009predicting} analyze linkage probability only from local topical graph structure. Further work like Weisfeiler-Lehman Neural Machine \cite{zhang2017weisfeiler} and LGNN \cite{zhang2018link} believe that it is sufficient to compute link likelihood only from the local neighborhood of a node pair and solve this task via various neural networks, GKC \cite{yuan2019graph} calculate the graph kernel similarities between subgraphs and use an SVM to decide each linkage. In general, there have been a lot of linkage prediction related work proposed and verified in the field of social networks, but there are relatively few works related to face recognition and clustering.

\section{Methodology}

In large-scale face clustering, supervised approaches
demonstrate their effectiveness with various mechanisms. Some supervised approaches\cite{yang2019learning} can handle complex patterns of clusters, but rely on manual components and large number of overlapping sub-graphs, which is time-consuming. Other light-GCN-based methods \cite{wang2019linkage, yang2020learning, guo2020density} can improve the speed of clustering, but are unable to consider both node and edge representations at the same time. To address the problem, we propose a simple yet efficient model: Face Transformer (FaceT). In FaceT, we divide the clustering task into two steps. First, we apply Relation Encoder (RE) to fuse contextual information for a node with original features from its neighbors. Then, Linkage Predictor (LP) is designed to determine whether paired nodes belong to the same ID or not. 


\subsection{Overview}

FaceT aims to generate accurate paired linkage predictions using two well-designed modules. Given a dataset $\mathcal{D}$, we extract deep features of images by a pretrained DCNN model. Let $\mathcal{F} = \{f_i\}_{i=1}^N$ as feature set where $f_i \in \mathbb{R}^D$, $D$ denotes the dimension of each image and $N$ denotes the number of images. For each sample feature $f_q$, we find its $hop_1$ nearest neighbor nodes by comparing their features similarities in $\mathcal{F}$, which are regarded as the candidate samples for final possible linkages. For query feature $f_q$ and those $hop1$ neighbors, we utilize the same manner to search the $hop_2$ nearest neighbor nodes for $f_q$ and its candidate samples, respectively, which is a scalable schema. Then we replace query feature $f_q$ with enhanced feature $g_q$ generated by RE. For all $hop1$ candidates, we apply the same manner and generate enhanced candidate features. As the enhanced query feature and its enhanced neighbors are generated, the LP is designed to assign linkage probabilities between query sample and its $hop1$ candidate samples. Finally, the obtained probability is used as the similarity score to determine if the candidate is connected to this query pivot. Given a threshold $\tau$, we form the link between query sample and a candidate sample whose pair linkage probability is larger than $\tau$. By repeating the above process while treating all samples as query samples, we can get the linkage predictions on the entire test set. Finally, clustering procedure with linkage predictions could be done with Union-Find algorithm \cite{fredman1989cell} 

The key challenge for the proposed method remains in how to aggregate local context into node embeddings and edge embeddings. As shown in Figure \ref{fig2}, our framework consists of two learnable modules, namely Relation Encoder(RE) and Linkage Predictor(LP). The former module aggregates local context information from its neighbors, and the latter module judges whether a pair of images belong to the same cluster or not. 

\subsection{Transformer Preliminary}
\label{sec:transformer_preliminary}

Assume we have $n$ query vectors each with dimension $d_q: Q \in \mathbb{R}^{n * d_q} $ , we can define an attention function $\mathcal{A}(Q, K, V)$ to calculate similarity between instance pairs. 

\begin{equation}
\mathcal{A}(Q,K,V; \omega) = \omega(QK^\top)V
\end{equation}

There $K \in \mathbb{R}^{n_v*d_1}$, $V \in \mathbb{R}^{n_v * d_v}$ are $n_v$ key-value pairs, $\omega$ is an activation function, the output $\omega(Q K^\top)V$ is a weighted sum of value vectors, where an instance value would get more weight when its key has larger dot product with the query vector. For multi-head attention $\mathcal{M}(\cdot, \cdot, \cdot; \lambda)$, first project $Q, K, V$ onto $h$ different $d_q^M$, $d_k^M$, $d_v^M$-dimensional vectors, then apply an attention function $\mathcal{A}(\cdot ; \omega_j)$ to each of these $h$ projections, finally reduce dimension with a linear transformation as follows:

\begin{equation}
\mathcal{M}(Q, K, V; \lambda, \omega) = cat(O_1, ..., O_h)W^O
\end{equation}

where,
\begin{equation}
O_i = \mathcal{A}(QW^Q_i, KW^K_i, VW^V_j; \omega_i)
\end{equation}

There $cat(\cdot, \cdot)$ represents concatenation operation along the feature dimension. Note that $\mathcal{M}(\cdot, \cdot, \cdot; \lambda)$ has learnable parameters $\lambda = \{W_i^Q, W_i^K, W_i^V\}^h_{i=1}$, where $W_i^Q, W_i^K \in \mathbb{R}^{d_q * d^M_q}$, $W^V_i \in \mathbb{R}^{d^v * d^M_v}$, $W^O \in \mathbb{R}^{hd^M_v*d}$. Unless otherwise specified, we use a scaled softmax $\omega_i(\cdot) = softmax(\frac{\cdot}{\sqrt{d}})$.


\subsection{Relation Encoder}

Relation Encoder(RE) is used for enhancing node representations, which consists of $d_e$ self-attention layers and one normal attention layer. Its structure is illustrated in figure \ref{fig:graphics3:a}. For each node embedding $f_q$ to be enhanced, first we find its $hop2$ neighbors $f_{q,1}, f_{q,2}, ..., f_{q, hop2}$ from $\mathcal{F}$ with normal retrieval methods. Then, we use self-attention layers to enhance $f_{q, j}$ , where $j \in \{1, 2, ..., hop2\}$ . 

Apart from the attention functions mentioned in \ref{sec:transformer_preliminary}, we apply dropout to each sub-layers output before it is added to the sub-layer input and normalized. We employ a residual connection around each of the two sub-layers, followed by layer normalization. In following method, dropout function is represented as $d(\cdot)$, layer normalization function is represented as $\mathcal{N}_L(\cdot)$. First, we use self-attention layers to enhance original face features, which can be formulated as:

\begin{figure}[tp]
  \subfigure[Relation Encoder]{
    \begin{minipage}{3.2cm}
    \includegraphics[scale=0.3]{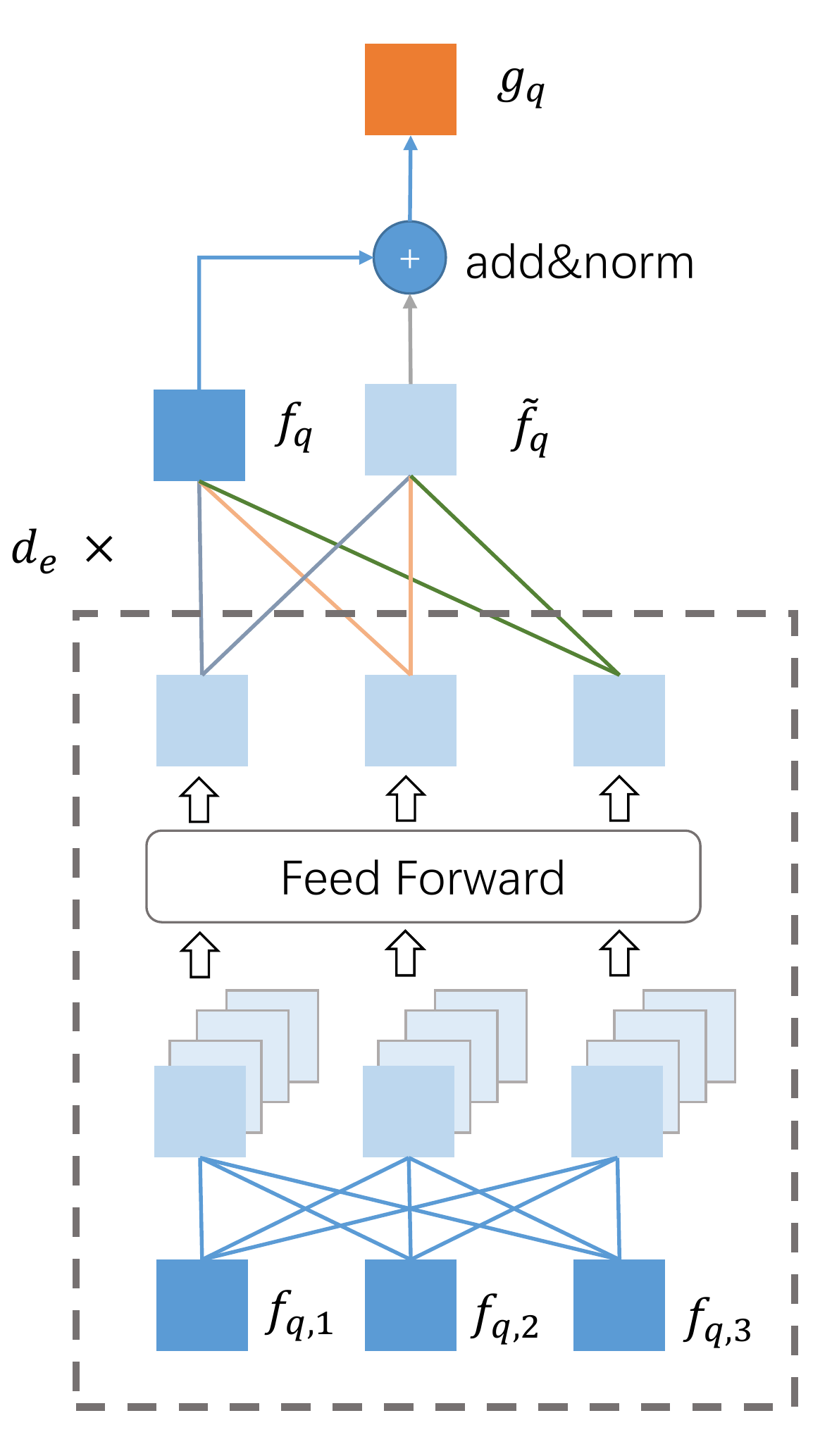}
    \label{fig:graphics3:a}
\end{minipage}
   }
   \hspace{1.5em}
  \subfigure[Linkage Predictor]{
    \begin{minipage}{3cm}
    \includegraphics[scale=0.3]{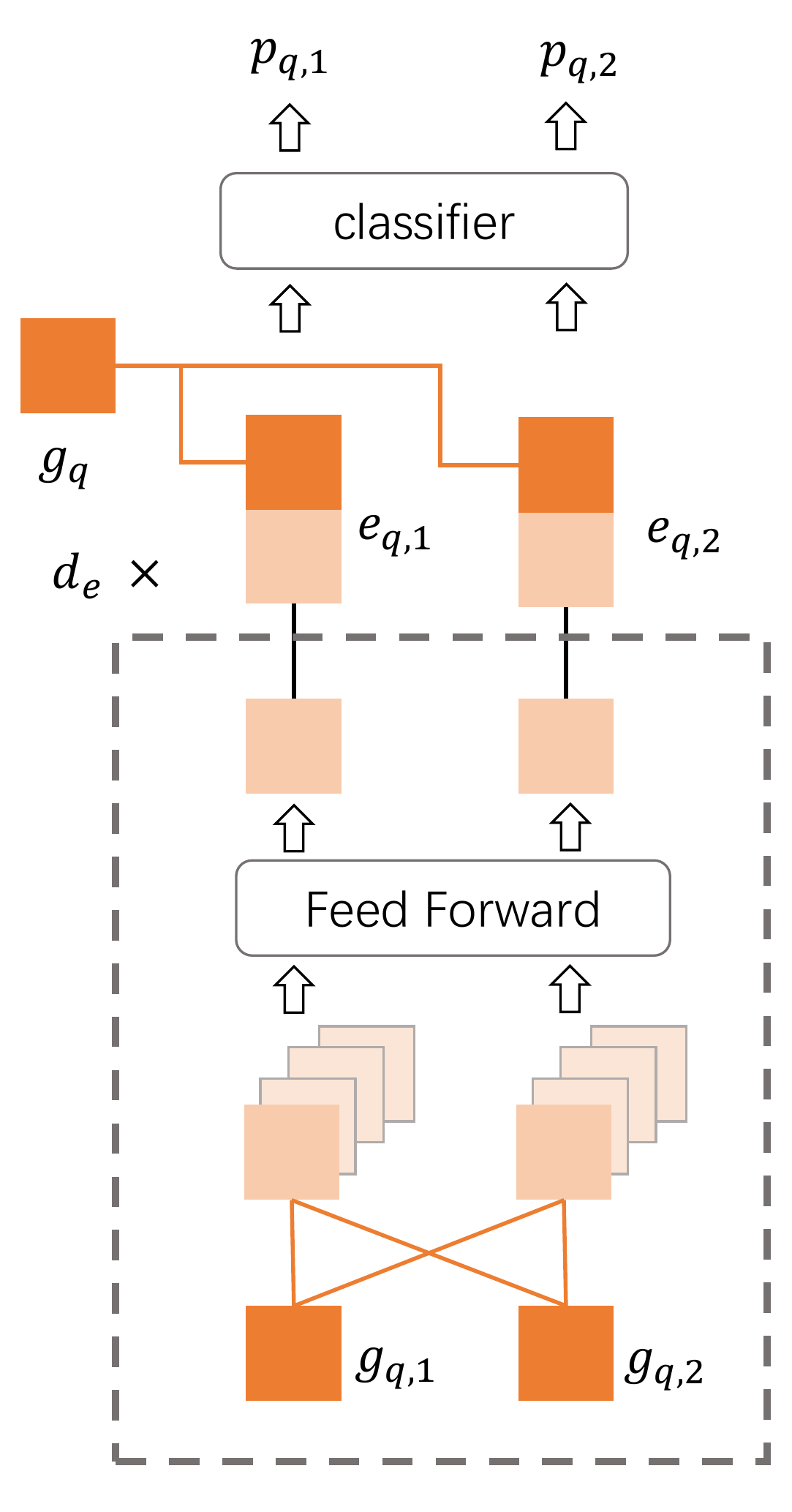}
    \label{fig:graphics3:b}
\end{minipage}
    }
  \caption{Details of Relation Encoder and Linkage Predictor.}%
    \label{fig3}
\end{figure} 



\begin{equation}
f^l_{q, k} = \mathcal{N}_L(d(\mathcal{M}(f^{l-1}_{q, k}, K^{l-1}, V^{l-1})) + f^{l-1}_{q, k})
\end{equation}

There $f^l_{q, k}$ denotes self-attention layer output from the $l-th$ layer, $l \in \{1, 2, ..., d_e\}$, $f^0_{q, k} = f_{q, k}$, $K, V$ denotes all $hop2$ neighbors' key vectors and value vectors. Next, we will use a common attention layer to construct the query's context representation, denotes as $\tilde{f_q}$ . Finally, we use an add and norm operation to generate enhanced face feature $g_k$, formulated as:

\begin{equation}
\tilde{f_q} = (\mathcal{M}(f_q, K^{d_e}, V^{d_e}))
\end{equation}

\begin{equation}
g_{q} = \mathcal{N}(\tilde{f_q} + f_q)
\end{equation}

\subsection{Linkage Predictor}

Linkage Predictor(LP) is used for enhancing edge representations and predicting linkages, which consists of $d_e$ self-attention layers and one MLP working as a linkage classifier. Its structure is illustrated in figure \ref{fig:graphics3:b}. This module is applied to decide whether a neighbors' enhanced face feature $g_{q, k}$ could build a connection with the enhanced query face feature $g_{q}$. First, we find $hop1$ nearest neighbors of query feature $f_q$ from $\mathcal{F}$ and replace them with related enhanced face features generated by RE. Secondly, self-attention layers are applied to enhanced neighbors' features, get $g^l_{k}$, formulated as:

\begin{equation}
g^l_{q, k} = \mathcal{N}_L(d(\mathcal{M}(g^{l-1}_{q, k}, K, V)) + g^{l-1}_{q, k})
\end{equation}

There $g^l_{q, k}$ denotes self-attention layer output from the $l-th$ layer, $l \in \{1, 2, ..., d_e\}$, specially, $g^0_{q, k} = g_{q, k}$, $K, V$ denotes all $hop1$ neighbors' key vectors and value vectors. Then, concatenate $g_{q}$ and $g^l_{q, k}$ to generate edge embedding $e_{q, k}$, defined as:

\begin{equation}
e_{q, k} = cat(g_k, g^{e_d}_{q, k})
\end{equation}

Finally, use a 2-layer MLP to predict probability $p_{q, k}$ of the $e_{q, k}$, which is formulated as: 

\begin{equation}
p_{q, k} = \omega(\mathcal{C}(e_{q, k}))
\end{equation}

There $\mathcal{C}(\cdot)$ is a two-layer MLP using an edge embedding $e_{q, k}$ as input, output probability $p_{q, k}$ of the input edge is positive. We use PReLU \cite{he2015delving} as activation function in our method.


\subsection{Complexity Analysis}

For both RE and LP, we need to construct a KNN graph; this needs to be done only once. Hence, its complexity can be regarded as $O(n log n)$ by Approximate Nearest Neighbor (ANN) search. Like the previous work LGCN, the proposed method is only processed on the local topical subgraph. Hence the runtime of the link prediction process grows linearly with the number of data. Therefore, the total complexity of our pipeline is $O(n log n)$ considering ANN cost, $O(n)$ for ignoring ANN cost, $n$ remarks instance amount of $\mathcal{D}$.

\section{Experiments}

\subsection{Experimental Settings}

\textbf{Face Clustering. } MS-Celeb-1M \cite{guo2016ms} is a large-scale face recognition dataset consisting of 100K identities and 5.8M images. As previous work DS-GCN and VE-GCN did, we adopt the widely used annotations from ArcFace \cite{deng2019arcface} that contains 5.8M images from 86K classes. The cleaned dataset was randomly split into ten parts, with an almost equal number of identities. To make a fair comparison, we use the same data released by DS-GCN and VE-GCN \footnote{https://github.com/yl-1993/learn-to-cluster}, details of those datasets are listed in Table \ref{table_1}, subset 0 is adopted for training face recognition model and clustering model, the others is used for testing. We use ArcFace\cite{deng2019arcface} as the face representations with dimension of 256.

Besides, we tested our method with additional DCNN model to validate the scalability of our method. We use ArcFace\cite{deng2019arcface} as the face representations with dimension of 512. This model is trained on the union set of MS-Celeb-1M \cite{guo2016ms} and VGGFace2\cite{2017VGGFace2} dataset. For a fair comparison, we use the data provided by LGCN\cite{wang2019linkage} \footnote{https://github.com/Zhongdao/gcn\_clustering}. Our FaceT is trained on CASIA dataset\cite{2014Learning} and tested on IJB-B dataset\cite{2017IARPA}. IJB-B consists of three sets, which include 512, 1,024, 1,845 identities, and 18,171, 36,575, 68,195 samples. We follow its official protocol\cite{2017IARPA} for evaluation. 

\textbf{Fashion Clustering.}
We also evaluate the effectiveness of our method over the non-face dataset. We adopted the large subset of DeepFashion \cite{liu2016deepfashion} , a long-tail clothes retrieval dataset. To make a fair comparison, we use the same data released by DS-GCN and VE-GCN \textsuperscript{\rm 1}. Training features and testing features are mixed in the original split and randomly sample 25, 752 images from 3, 997 categories for training, and the other 26, 960 images with 3, 984 categories for testing. Because fashion clustering is regarded as an open set problem, there is no overlap between training categories and testing categories. 

\begin{table}[t]
\small
\renewcommand\arraystretch{1.0}
  \begin{center}
  \caption{Details of MS-Celeb-1M subsets.}
\label{table_1}
 \begin{tabular}{@{}cccccccccccccccc@{}}
   \toprule
     id & 0 & 1 & 3 & 5 & 7 & 9 \\
    \hline
     \#id & 8.6K & 8.6K & 25.7K & 42.9K & 60.0K & 77.2K \\
     \#inst & 576K & 584K & 1.74M & 2.89M & 4.05M & 5.21M \\
     \bottomrule
    \hline
  \end{tabular}
  \end{center}
\vspace{-1pt}
\end{table}

\textbf{Evaluation Metrics.}
To evaluate the performance of the proposed clustering algorithm, we adopt three mainstream evaluation metrics: BCubed Fmeasure \cite{amigo2009comparison}, pairwise Fmeasure \cite{banerjee2005model} and normalized mutual information(NMI). 

NMI is a widely used metric that measures the normalized similarity two sets, given $\Omega$ the ground truth cluster set, $C$ the prediction cluster set, $H(\Omega)$ and $H(C)$ are their entropies, $I(\Omega, C)$ represents the mutual information. NMI is calculated as follows.

\begin{equation}
NMI(\Omega, C) = \frac{I(\Omega, C)}{\sqrt{H(\Omega)H(C)}}
\end{equation}

We did not evaluate traditional methods with the NMI measure metric on the MS1M benchmark. For the IJB-B benchmark, $F_{512}, F_{1024}, F_{1845}$ are Bcubed F-scores of different sets.

\textbf{Implementation Details.}
To train the FaceT, we set the depth of encoders $d_e$ as 2, both for RE and LP. Our feature dimension is 256, attention head dimension is 64, attention head amount is 4, and the dropout ratio is 0.4. For IJB-B experiments, our feature dimension is 512,  attention head dimension is 128, the other hyper parameter settings remain the same.


We adopt a linear learning rate warm-up for the first 500 steps for the training phase, then use cosine decay policy for the rest training epochs. The batch size is 32; the weight decay parameter is 0.0005. For MS-Celeb-1M and CASIA dataset, we adopt $hop1=150, hop2=5$, training for 60 epochs with a base learning rate 0.002; for DeepFashion dataset, we use $hop1=8, hop2=6$, training for 1200 epochs with a base learning rate 0.02.

\begin{table*}[hbpt]
\small
\renewcommand\arraystretch{1.0}
  \begin{center}
  \caption{Comparison on MS-Celeb-1M with different numbers of unlabeled images, subset id remarks different sizes. 1,3,5,7,9 respectively represent images count of 584K, 1.74M, 2.89M, 4.05M and 5.21M .}
\label{table_2}
  \setlength{\tabcolsep}{4.25pt}
  \begin{tabular}{@{}cccccccccccccccc|l@{}}
   \toprule
     Method & \multicolumn{5}{c|}{pairwise F-score($F_P$)} & \multicolumn{5}{c}{BCubed F-score($F_B$)} & \multicolumn{5}{|c}{NMI} & \multicolumn{1}{|l}{Time} \\
    \cline{1-16}
    subset id & 1 & 3 & 5 & 7 & 9 & 1 & 3 & 5 & 7 & 9 & 1 & 3 & 5 & 7 & 9 &  \\
    \hline
     K-Means & 79.21 & 73.04 & 69.83 & 67.9 & 66.47  & 81.23 & 75.2 & 72.34 & 70.57 & 69.42 & - & - & -  & - & - & 11.5h \\
     HAC & 70.63 & 54.4 & 11.08 & 1.4 & 0.37 & 70.46 & 69.53 & 68.62 & 67.69 & 66.96 & - & - & -  & - & - & 12.7h \\
     DBSCAN & 67.93 & 63.41 & 52.5 & 45.24 & 44.94 & 67.17 & 66.53 & 66.26 & 44.87 & 44.74 & - & - & -  & - & - & \textbf{1.9m} \\
     ARO & 13.6 & 8.78 & 7.3 & 6.86 & 6.35 & 17 & 12.42 & 10.96 & 10.5 & 10.01 & - & - & -  & - & - & 27.5m \\
    \hline          
     CDP & 75.02 & 70.75 & 69.51 & 68.62 & 68.06 & 78.70 & 75.82 & 74.58 & 73.62 & 72.92  & 94.69 & 94.62 & 94.63 & 94.62 & 94.61 & 2.3m \\
     L-GCN & 78.68 & 75.83 & 74.29 & 73.70 & 72.99 & 84.37 & 81.61 & 80.11 & 79.33 & 78.60 & 96.12 & 95.78 & 95.63 & 95.57 & 95.49 & 63.8m \\
     DS-GCN & 87.61 & 83.76 & 81.62 & 80.33 & 79.21 & 87.76 & 83.99 & 82.00 & 80.72 & 79.71 & 97.04 & 96.55 & 96.33 & 96.18 & 96.07 & 47.3m \\
     VE-GCN & 87.93 & 84.04 & 82.1 & 80.45 & 79.30 & 86.09 & 82.84 & 81.24 & 80.09 & 79.25 & 96.41 & 96.03 & 95.85 & 95.71 & 95.62 & 11.5m \\
    \hline 
     \textbf{FaceT} & \textbf{91.12} & \textbf{89.07} & \textbf{86.78} & \textbf{84.10} & \textbf{83.86}  & \textbf{90.50} & \textbf{86.84} & \textbf{85.09} & \textbf{84.67} & \textbf{83.86} & \textbf{97.61} & \textbf{97.12} & \textbf{96.87} & \textbf{96.82} & \textbf{96.67} & 28.0m \\
    \bottomrule
  \end{tabular}
  \end{center}
\vspace{-10pt}  

\end{table*}

\begin{table}[t]
\small
\renewcommand\arraystretch{1.0}
  \begin{center}
  \caption{Comparison on IJB-B. }
    \label{table_3}
  \setlength{\tabcolsep}{15pt}
  \begin{tabular}{@{}cccccccccccccccc@{}}
   \toprule
     Method & $F_{512}$ & $F_{1024}$ & $F_{1845}$ \\
    \hline
     K-means & 61.2 & 60.3 & 60.0 \\
     DBSCAN & 75.3 & 72.5 & 69.5 \\
     Spectral & 51.7 & 50.8 & 51.6 \\
     AP & 49.4 & 48.4 & 47.7 \\
     ARO & 76.3 & 75.8 & 75.5 \\
     DDC & 80.2 & 80.5 & 80.0\\
     LGCN & 83.3 & 83.3 & 81.4 \\
    \hline
     \textbf{FaceT} & 83.1 & 83.3 & 82.2 \\
     \bottomrule
  \end{tabular}
  \end{center}
\vspace{-10pt}  
\end{table}

\begin{table}[t]
\small
\renewcommand\arraystretch{1.0}
  \begin{center}
  \caption{Comparison on DeepFashion. }
\label{table_4}
  \setlength{\tabcolsep}{15pt}
  \begin{tabular}{@{}cccccccccccccccc@{}}
   \toprule
     Method & $F_P$ & $F_B$ & NMI \\
    \hline
     K-means & 32.02 & 53.30 & 88.91 \\
     HAC & 22.54 & 48.77 & 90.44 \\
     DBSCAN & 25.07 & 53.23 & 90.75 \\
     MeanShift & 31.61 & 56.73	& 89.29 \\
     Spectral & 29.60 & 47.12 & 86.95 \\
     ARO & 25.04 & 52.77 & 88.71 \\
     CDP & 28.28 & 57.83 & 90.93 \\
     LGCN & 30.7 & 60.13 & 90.67 \\
     DS-GCN & 33.25 & 56.83 & 89.36 \\
     VE-GCN & \textbf{38.47} & 60.06 & 90.50 \\
    \hline
     \textbf{FaceT} & 34.82 & \textbf{61.29} & \textbf{91.28} \\
     \bottomrule
  \end{tabular}
  \end{center}
\vspace{-10pt}  

\end{table}

\subsection{Method Comparison}

Like the previous works, we compare our method with several baseline methods, each with a brief description. For all those methods, we tune the hyper-parameters and report the best results.

\noindent \textbf{K-Means} \cite{lloyd1982least} is a commonly used clustering algorithm.
For N $\geq$ 1.74M, we use mini-batch K-means \cite{sculley2010web}, which can reduce running time by running the original K-Means algorithm with mini-batches.

\noindent \textbf{HAC} \cite{Sibson1973SLINK} hierarchically merges close clusters based on various criteria in a bottom-up manner.

\noindent \textbf{DBSCAN} \cite{ester1996density} recognize clusters from the gallery based on a designed density criterion, then leave the isolated point.

\noindent \textbf{MeanShift} \cite{cheng1995mean} updates candidates for centroids to be the mean of the points within a given region.

\noindent \textbf{AP} \cite{Frey2007Clustering} creates clusters by sending messages between pairs of samples until convergence.

\noindent \textbf{Spectral Clustering} \cite{shi2000normalized} performs a low-dimension embedding of the affinity matrix between samples, followed by clustering of the components of the eigenvectors in the low dimensional space. 

\noindent \textbf{DDC} \cite{lin2018deep} performs clustering  based on measuring density affinities between local neighborhoods in the feature space.

\noindent \textbf{ARO} \cite{otto2017clustering} performs clustering with an approximate nearest neighbor search and a modified distance measure.

\noindent \textbf{CDP} \cite{zhan2018consensus} performs clustering by exploiting a more robust pairwise relationship via gathering different predictions.

\noindent \textbf{LGCN} \cite{wang2019linkage} is a supervised clustering method that adopts GCNs to exploit graph context for pairwise prediction, then performs clustering based on the output pairwise prediction.

\noindent \textbf{DS-GCN} \cite{yang2019learning} is a supervised clustering method that formulates clustering as a detection and segmentation pipeline.

\noindent \textbf{VE-GCN} \cite{yang2020learning} proposes a method that employs a stacked GCN architecture to estimate the connectivity and obtain clusters by connecting each node to the most connective neighbors in the candidate set.


\noindent \textbf{FaceT} is the proposed method that uses a hierarchical transformer architecture to enhance node embedding and edge embedding synchronously and build connections on enhanced pairwise linkage predictions.

\subsection{Results}
For all methods, we tune the corresponding hyper-parameters and report the best results. The results in Table \ref{table_2}, Table \ref{table_3}, and Table \ref{table_4} show:
(1) As is conducted in \cite{yang2019learning, yang2020learning}, traditional methods including K-Means, DBSCAN, Spectral and ARO are limited in number of clusters assumptions, distribution assumptions or large computational budget, which make them hard to use in real-world situations.
(2) CDP is quite efficient and achieves considerable F-scores on different datasets.
(3) L-GCN and DS-GCN surpasses CDP consistently but they are an order of magnitude slower than CDP. 
(4) VE-GCN yields superior performance, and has a high operating efficiency benefit from its high-confidence sample screening strategy.
(5) The proposed FaceT outperforms previous methods consistently. Although the training set of FaceT only contains 584K images \ref{table_2}, it scales well to 5.21M unlabeled data, demonstrating its effectiveness in capturing contextual information of nodes and edges and predicting linkages.

\begin{figure}[t]
\centering
\includegraphics[width=0.9\columnwidth]{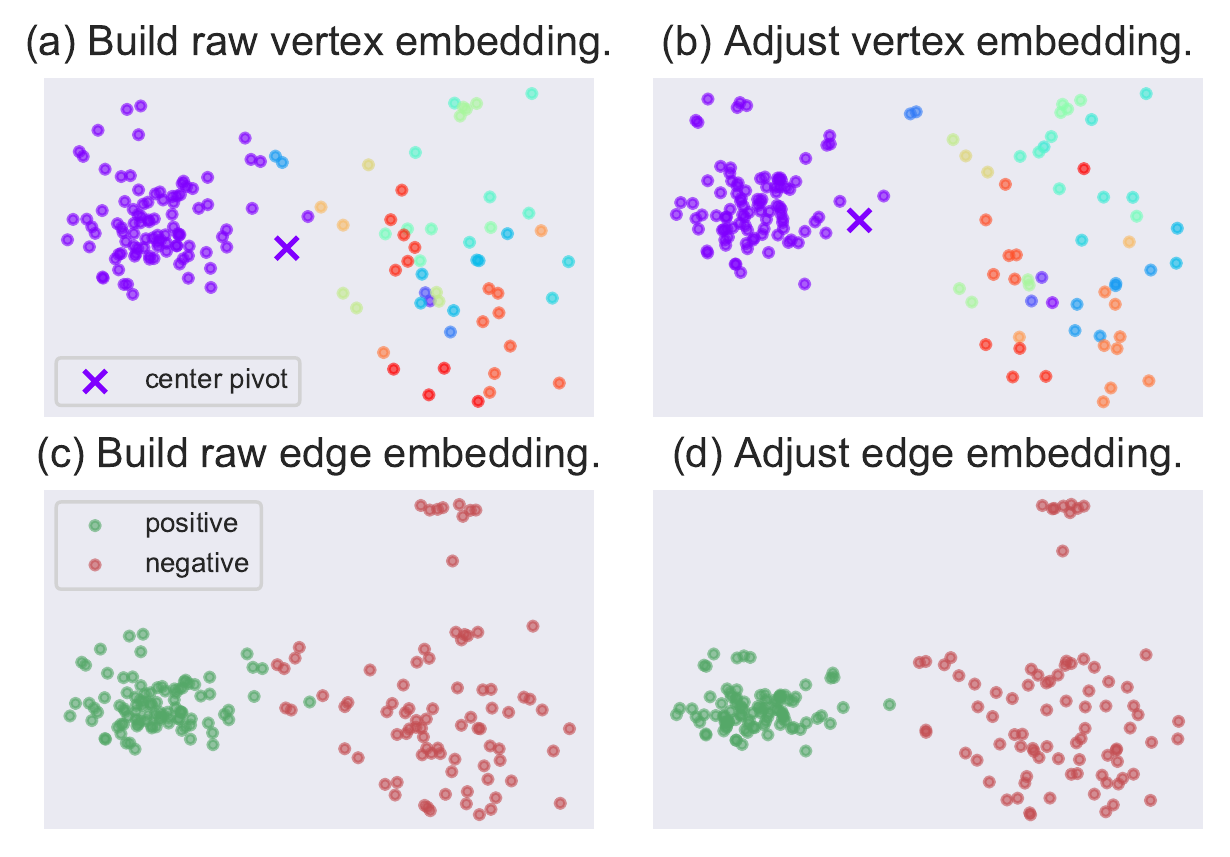} 
\caption{Adjust node and edge embedding with our method.
(a) Original features as node embeddings, where samples from the same category are marked in the same color. (b) Enhanced node feature embeddings generated by RE. (c) Raw edge embeddings by concatenating paired node features. (d) Enhanced edge feature embeddings generated by LP. 
Our method makes node of the same class concentrate more tightly; meanwhile, the positive and negative edge embedding interface is more explicit. This affinity graph is built on a random center pivot's local topical subgraph of MS1M dataset subset 1. }
\label{fig4}
\end{figure}

For results on IJB-B in Table \ref{table_4},  L-GCN\cite{wang2019linkage} uses the down-sampled CASIA training set, while sampling details is not provided, we randomly make three down-sampled subsets for training and take the best result to report. Therefore, comparison on IJB-B benchmark is not fair enough to make a solid conclusion due to the unaligned down-sampling policy, and we should focus on datasets that can be fairly compared, such as MS1M \ref{table_2}. What's more, DA-NET\cite{guo2020density} used completely different feature and training set, therefore it is hard to make a fair comparison between DA-NET and the other methods.


\subsection{Ablation Study}
In order to verify that key modules work as expected and study some key design choices, we conduct ablation study on MS-Celeb-1M subset 1.
The ablation experiments is conducted from two aspects, the validity of the model structure and the influence of model hyper-parameters. 

\textbf{Visual Analysis.}
Regarding the impact of our method, we give a visual analysis in Figure \ref{fig4}. We randomly selected the center point of a batch and its neighbors of hop1, and visualized the point embedding and edge embedding.  As is shown in Figure \ref{fig4}, after adjustment by Face Transformer, the sample variance of the same category as the center point becomes smaller, and the boundary between positive and negative samples on the edge embedding interface becomes clearer, which means those ambiguous samples are easier to be separated after those two adjustments.

\textbf{Relation Encoder and Linkage Predictor}.
To better evaluate each sub structure's role, we conduct end-to-end experiments using each sub-module separately. Because our method is composed of RE and LP structures in series, using these two sub-modules independently requires separate retraining. For validating the effectiveness of LP, we trained a model with only one Relation Encoder structure, which is termed as \textbf{only RE}. For validating the effectiveness of RE, we trained a model with only one Linkage Predictor structure, which is termed as \textbf{only LP}. Training policy and hyper-parameters are the same as the original FaceT. An end-to-end clustering test is shown in Table \ref{table_5}.

\begin{figure}[t]
\centering
\includegraphics[width=0.9\columnwidth]{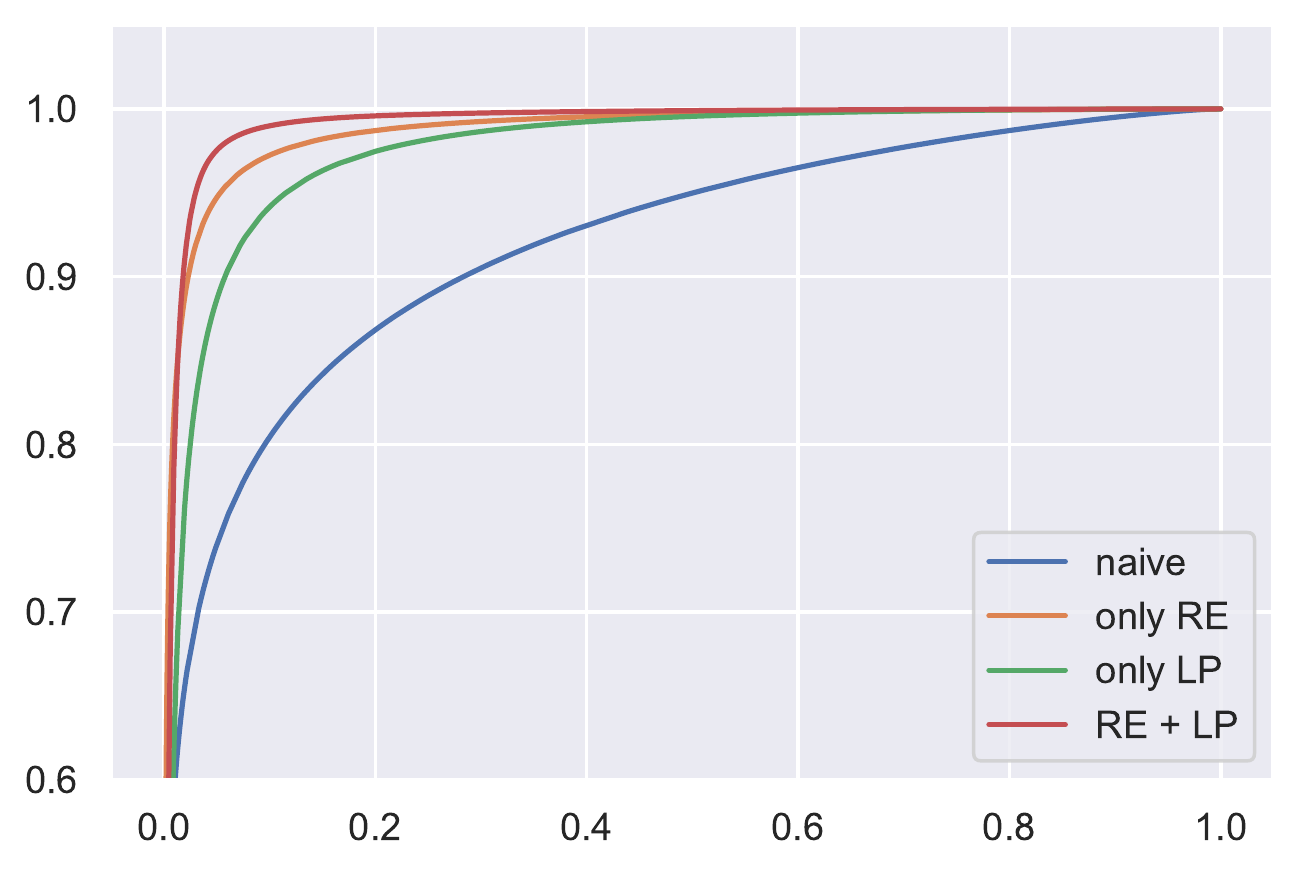} 
\caption{ROC of Linkage prediction on MS-Celeb-1M.}
\label{fig5}
\end{figure}

\begin{table}[t]
\small
\caption{Compatibility analysis on MS-Celeb-1M.}
\renewcommand\arraystretch{1.0}
  \begin{center}
  \setlength{\tabcolsep}{17.5pt}
 \begin{tabular}{@{}cccccccccccccccc@{}}
   \toprule
     Method & $F_P$ & $F_B$ & NMI \\
    \hline
     naive & 70.63 & 70.46 & 92.67 \\
     only RE & 86.17 & 87.29 & 96.72 \\
     only LP & 82.08 & 80.72 & 94.72 \\
    \hline
     \textbf{RE+LP} & \textbf{91.12} & \textbf{90.50} & \textbf{97.61} \\
    \bottomrule
  \end{tabular}
  \end{center}
\vspace{-10pt}

\label{table_5}
\end{table}

In order to properly train \textbf{only RE} model, we need to introduce a new training objective function to complete end-to-end training. We first normalize of node features with L2 normalize function $\mathcal{N}$ , then use the L2 distance as the distance metric between the two nodes, finally convert this distance metric into a two-class probability distribution to use the same training strategy as the MLP-based two-classifier. The specific method could be described as follows.

\begin{equation}
e_{q, k} = 0.25 * ||\mathcal{N}(g_k) - \mathcal{N}(g_{q, k})||^2_2
\end{equation}

\begin{equation}
p_{q, k} = (1 - e_{q, k}, e_{q, k})
\end{equation}

In the end-to-end test scenario, the complete FaceT architecture has obvious performance advantages. In the case of limited computing resources, using RE or LP alone can also achieve a particular effect. Additionally, we utilize the ROC curve to illustrate the discriminative power of our final linkage prediction approach. There, we use the top 80 predictions of each node on MS-Celeb-1M subset 1. As shown in Figure \ref{fig5}, under the same false-positive rate, the full FaceT architecture's true-positive rate is much higher than the other candidates.

\textbf{Time Complexity Analysis}.
We measure the runtime of different methods with ES-2640 v3 CPU and Tesla V100 (16G). For MS1M, we measure the clock timings when N = 584K. (i.e. subset 1) Note that the clustering framework we used is similar to LGCN, which is based on the clustering task completed by linkage prediction and clustering post-processing, and it is quite different from DS-GCN and VE-GCN in clustering framework. The existing clustering methods have large differences in specific implementation and resource utilization, so it is not completely fair to directly compare the running time. Because our CPU environment is consistent, we directly use the speed measurement results of some methods in \cite{yang2020learning}.

\textbf{Hyper-parameters}.
To better understand our hyper-parameters, we set a group of parameters as our default setting, then alter one parameter while the others are fixed. We mainly study the influence of encoder depth ($d_e$), attention head amount($n_h$), attention head dimension($d_h$), dropout ratio($dr$), training hop1($hop1$) and training hop2($hop2$). For the default method, ($d_e$, $n_h$, $d_h$, dropout, $hop1$, $hop2$) is set as (2, 4, 64, 0.4, 150, 5). 


\begin{table}[hbpt]
\small
\renewcommand\arraystretch{1.0}
  \begin{center}
  \caption{Hyper parameter analysis. We alter only one hyper parameter to classify its robustness each time,
$F_P$, $F_B$, NMI are different measure metrics.}
\label{table_6}
  \setlength{\tabcolsep}{12.5pt}
 \begin{tabular}{@{}cccccccccccccccc@{}}
   \toprule
      & value & $F_P$ & $F_B$ & NMI  \\
    \hline
     default & - & 91.12	& 90.50 & 97.61 \\
    \hline
     $d_e$ & 1 & 91.08 & 90.46 & 97.56 \\
      & 4 & 91.07 & 90.39 & 97.57 \\
    \hline
     $n_h$ & 2 & 91.12 & 90.37 & 97.47 \\
       & 8 & 91.34 & 90.6 & 97.63 \\
    \hline
     $d_h$ & 32 & 90.96 & 90.18 & 97.18 \\
       & 128 & 91.07 & 90.6 & 97.63\\
    \hline
     $dr$ & 0.2 & 91.64 & 90.75 & 97.67 \\
       & 0.6 & 89.96 & 89.96 & 97.47 \\
    \hline
     $hop1$ & 120 & 91.18 & 90.39 & 97.58\\
       & 180 & 91.51 & 90.54 & 97.63 \\
    \hline
     $hop2$ & 10 & 90.76 & 90.26 & 97.56 \\
      & 20 & 89.03 & 89.23 & 97.29 \\
    \bottomrule
  \end{tabular}
  \end{center}
\vspace{-10pt}  

\end{table}


From the listed experimental results in Table \ref{table_6}, our method is not sensitive to the encoder depth $d_e$. From our experimental results, reducing the numbers of self-attention layers in RE and LP will have a certain impact on the end-to-end test results. We believe that the self-attention layers can make the neighborhood topology between nodes smoother, thereby making the normal attention layer of RE and the MLP classifier of LP generate more accurate results. However, under the condition that other parameters are fixed, deepening the encoders will not bring substantial benefits. We believe that when $d_e$ is set to 2, the model's fitting ability is already sufficient to learn the context distribution. 

For the attention head related parameters, such as $n_h$ and $d_h$, increasing the parameter scale can bring certain degree of performance improvement. It is a typical strategy of increasing the amount of calculation in exchange for performance improvement. However, it is observed that this improvement is not prominent, and trade-off should be decided on specific actual application scenarios.

For training hyper-parameter $dr$, we found that 0.4 used in the default settings may be too large. Broad regularization terms may limit the neural network model's learning ability, which may lead to under-fitting. From the listed experimental results,  reducing the dropout ratio can slightly improve the performance on the test set.

For the heuristic parameters $hop1$ and $hop2$, the selected strategy is closely related to the data set. For $hop1$ neighbors, we expect the positive and negative samples to be balanced as much as possible while trying to make the model thoroughly learn some knowledge of difficult samples (i.e., positive samples with low similarity and negative samples with high similarity). During the training process, we found that loss tends to converge to a lower solution when hop1 is set to a small value, but it is difficult to thoroughly learn the knowledge of difficult samples, which leads to worse performance on the test set. For the nearest neighbors of $hop2$, we want them to come from the same category as the center point as much as possible, so this value is often set to a small value. Similar to the $hop1$ hyperparameter, during the training process, we found that increasing this hyper-parameter leads to a smaller loss converge, but performs worse on the test set. We think this phenomenon is caused by overfitting. When the number of neighbor samples used for training RE increases, a trivial solution may be learned from training samples that not from the same category of center pivot, limiting the performance of the model on the test set. 

\section{Conclusion and Future Work}
In this work, we refer to the idea of Transformer applied in document classification and model the clustering task as a linkage prediction task. This paper proposes a novel method to enhance the node and the edge representation simultaneously and achieve the most advanced performance on the face clustering task. Experiments on clustering scenarios have verified the effectiveness of the method. Besides, we notice that using RE alone obtains robust node representations that are measurable in Euclidean space. As iterative updating policy of KNN may obtain more accurate results, the specific algorithm framework is worthy of further exploration. Finally, there is still room for further improvement in the current method of hyper-parameter exploration. We will make further theoretical analysis and experimental exploration in the model structure design and hyper-parameter settings.

{\small
\bibliographystyle{ieee_fullname}
\bibliography{egbib}
}

\end{document}